\documentclass[runningheads]{llncs}

\usepackage{eccv}

\usepackage{eccvabbrv}

\usepackage{graphicx}
\usepackage{booktabs}

\usepackage[accsupp]{axessibility}  

\usepackage[pagebackref,breaklinks,colorlinks,citecolor=eccvblue]{hyperref}

\usepackage{orcidlink}

\begin{document}

\title{Frequency Matters: Explaining Biases of Face Recognition in the Frequency Domain}

\titlerunning{Frequency Matters}

\author{Marco Huber\inst{1,2}\orcidlink{0000-0003-3413-6291} \and
Fadi Boutros\inst{1}\orcidlink{0000-0003-4516-9128} \and
Naser Damer\inst{1,2}\orcidlink{0000-0001-7910-7895}}

\authorrunning{M.~Huber et al.}

\institute{Fraunhofer Institute for Computer Graphics Research IGD, Darmstadt, Germany \and
Technical University of Darmstadt, Darmstadt, Germany}

\maketitle

\begin{abstract} 
Face recognition (FR) models are vulnerable to performance variations across demographic groups. The causes for these performance differences are unclear due to the highly complex deep learning-based structure of face recognition models. Several works aimed at exploring possible roots of gender and ethnicity bias, identifying semantic reasons such as hairstyle, make-up, or facial hair as possible sources. Motivated by recent discoveries of the importance of frequency patterns in convolutional neural networks, we explain bias in face recognition using state-of-the-art frequency-based explanations. Our extensive results show that different frequencies are important to FR models depending on the ethnicity of the samples.
\end{abstract}


\vspace{-5mm}
\section{Introduction}
\vspace{-3mm}
Over the past few years, face recognition (FR) technologies have been widely adopted and applied in various application scenarios, impacting the daily lives of a diverse set of users.
These users are of different genders, ethnicities, and ages, which poses a particular challenge for FR systems as they should guarantee the same usability and security regardless of a user's demographic. Several recent works have shown that FR systems are biased towards demographic (e.g. ethnicity, gender, or age) \cite{DBLP:conf/ciarp/AcienMVBF18, DBLP:conf/icb/AtzoriFM22} and non-demographic attributes (e.g. hairstyle, illumination, ...) \cite{9534882, DBLP:conf/euvip/HerranzGD22} leading to verification performance variations, discriminating certain groups. These performance disparities motivated several researchers to dive deeper into possible reasons and investigate sources of this performance bias. Identified possible source for different kind of bias so far are, for example, facial hair \cite{DBLP:conf/wacv/WuTBORB24}, hairstyle \cite{DBLP:conf/wacv/BhattaABK23}, illumination \cite{DBLP:conf/cvpr/WuAKKB23}, make-up \cite{DBLP:conf/wacv/AlbieroSVZKB20}, resolution \cite{DBLP:conf/wacv/BhattaPKB24}, and model behavior \cite{10581908, DBLP:conf/icb/FuD22}. All of these works share that they focus on the semantic indicators of the image when investigating possible sources of bias.

A possible reason why the sources of bias remain unclear is that state-of-the-art FR models heavily rely on highly complex deep learning-based models that lack transparency and do not provide an intuitive understanding of their inner workings \cite{DBLP:journals/corr/abs-2208-09500}. Since faces are usually provided as images, the usage of convolutional neural networks (CNNs) dominates current FR systems \cite{DBLP:journals/ijon/WangD21a}. These CNNs process the information present in the face image data and are trained to learn a discriminative identity representation. Once the model is trained, it can be used to extract identity representation from images for verification or identification tasks. To increase the understanding of the inner workings of these CNNs, several explainable FR approaches have been proposed recently. These approaches provide additional context in the form of a saliency or heat maps that highlight the important areas for a certain FR decision \cite{DBLP:conf/wacv/HuberLTD24, DBLP:conf/wacv/KnocheTHR23, DBLP:conf/cvpr/Mery22, DBLP:conf/wacv/MeryM22, DBLP:conf/biosig/BousninaACP23, DBLP:conf/biosig/CorreiaCP23, DBLP:conf/wacv/LuXE24, 10581925}.  


Previous works showed that CNNs are not only using visible "semantic" clues but also imperceptible frequency patterns when processing the input data \cite{DBLP:conf/cvpr/WangWHX20, DBLP:conf/cvpr/AbelloHW21}. Wang \etal \cite{DBLP:conf/cvpr/WangWHX20} investigated the relationship between frequency components and the generalization behavior of CNNs during the training process and concluded that CNNs capture high-frequency patterns that misalign with human visual preference. Abello \etal \cite{DBLP:conf/cvpr/AbelloHW21} proposed a framework to divide the frequency spectrum into equally sized disjoint discs to investigate the impact of different discs on the classification decision and their experiments did not indicate a model bias to high frequencies. Most recently for FR, \cite{huber2024spatialexplanationsexplainableface} proposed to utilize the frequency domain for explaining FR decisions. In their approach, certain frequencies of face images are masked in frequency space by removing any information present in the frequencies, and then the impact of the masking on the similarity score is observed. The observed impact is interpreted as the importance of the frequency to explain the verification decision. More details about the approach of \cite{huber2024spatialexplanationsexplainableface} is provided in Subsection \ref{meth}.

In this work, we investigate possible sources of ethnicity bias in FR beyond visual factors such as hairstyle or facial hair. We utilize the frequency-based explainability approach of \cite{huber2024spatialexplanationsexplainableface} in extensive experiments on two state-of-the-art models and on five trained models with varying ethnicity bias.  Our results show that based on the ethnicity of samples, different frequencies are important to the model, explaining bias beyond visual factors. With our experiments on biased models, we observed that these importance differences increase with higher bias, and frequency importance shifts can be observed when compared with a baseline model. 
\vspace{-4mm}
\section{Related Work}
\vspace{-3mm}
\subsection{Bias in Face Recognition}
\vspace{-3mm}
\label{subsec:bias}
The ISO/IEC DIS 19795-10 standard \cite{ISO} refers to bias in biometric decisions as the "differential performance" which is defined as the "difference in biometric systems metrics across different demographic groups". A common understanding is that bias refers to performance variations that are dependent on a particular sub-group (e.g. gender, ethnicity, etc.).

The majority of studies on bias in FR focused on investigating and mitigating the vulnerability of different demographic groups. Several works showed that the verification performance of FR systems is worse on female faces than on male faces \cite{DBLP:conf/icb/AlbieroZB20, DBLP:conf/bmvc/AlbieroB20, DBLP:conf/wacv/AlbieroSVZKB20, DBLP:conf/wacv/BhattaABK23, DBLP:conf/wacv/WuTBORB24} and several possible causes were investigated. Albiero \etal \cite{DBLP:conf/icb/AlbieroZB20} performed experiments on the impact of gender-unbalanced training data and concluded, based on their experimental results that balancing the training data does not necessarily reduce bias. In \cite{DBLP:conf/bmvc/AlbieroB20}, face size and the amount of face pixels occluded due to hairstyle have been identified as a cause of bias. Other studies made similar observations regarding the impact of hairstyle \cite{DBLP:conf/wacv/BhattaABK23} and facial hair \cite{DBLP:conf/wacv/WuTBORB24} on performance variations between female and male faces. The impact of make-up has been investigated in \cite{DBLP:conf/wacv/AlbieroSVZKB20} and has been identified as another source of gender bias. Performance variations can also be observed for different ethnicities \cite{DBLP:conf/ciarp/AcienMVBF18, DBLP:journals/tbbis/CavazosPCO21, DBLP:conf/cvpr/SVKAB19, DBLP:journals/tifs/KlareBKBJ12}. The most commonly used datasets in this area for training and evaluation, Racial Faces-in-the-Wild (RFW) \cite{DBLP:conf/iccv/WangDHTH19}, Balanced Faces in the Wild (BFW) \cite{DBLP:conf/cvpr/RobinsonLHQ0T20}, and BUPT \cite{DBLP:journals/pami/WangZD22, DBLP:conf/cvpr/WangD20} introduce four ethnicity subsets that individuals belong to: \textit{Indian}, \textit{Black/African}, \textit{Asian}, \textit{Caucasian/White}. This probably oversimplifies a complex, naturally non-binary attribute as individuals may belong or identify to more than one ethnicity. Other works therefore propose to rather use skin tone \cite{DBLP:conf/biosig/MolinaCT20}. In \cite{DBLP:conf/cvpr/WangD20}, Skewed Error Ratio (SER) has been introduced as a bias measuring metric. Wu \etal \cite{DBLP:conf/biosig/GalballyHB18} identified face illumination and face region brightness, which discriminates darker skin tones as a possible cause for ethnicity bias. Data imbalance has also been observed as a cause of bias \cite{DBLP:journals/pami/WangZD22, DBLP:conf/iccv/WangDHTH19}. To go beyond demographic bias, \cite{9534882} analyzed over 47 different demographic and non-demographic attributes for their influence on verification performance. This included attributes such as sunglasses, face geometry, or hair color.

Most recently, few works approached bias and causes of bias from a different perspective and utilized explainability methods to investigate biased model behavior \cite{DBLP:conf/icb/FuD22, 10581908}. \cite{DBLP:conf/icb/FuD22} applied Score-CAM \cite{DBLP:conf/cvpr/WangWDYZDMH20} on single face images of ethnicity-split face image datasets and analyzed statistical differences of the obtained attention maps to identify certain face regions that might cause different model behavior based on the ethnicity. In contrast to single face images in \cite{DBLP:conf/icb/FuD22}, \cite{10581908} considered face image pairs and the verification decision of an FR system as the point of bias investigation. They applied two explainable FR approaches on verification decisions of FR models with different biases and observed a wider distributed attention of biased models compared to a baseline model. 


The previously mentioned work so far focused on the visual investigation of possible causes of biased model behavior. With the recent insights into the importance and influence of frequency patterns for CNNs \cite{DBLP:conf/cvpr/WangWHX20, DBLP:conf/cvpr/AbelloHW21, huber2024spatialexplanationsexplainableface} we propose to explain ethnicity bias
by observing the impact of certain frequency bands on the similarity score based on the ethnicity. This analysis extends current investigations \cite{DBLP:conf/icb/FuD22, 10581908} as it also leverages human imperceptible features that are utilized by CNNs when processing different ethnicities and cannot be investigated in the spatial domain, due to their frequency pattern nature.

Finally, its important to note that bias is not limited to FR systems, but is also present in other biometric applications and scenarios such as presentation attack detection \cite{DBLP:conf/eusipco/FangDKK20, DBLP:journals/pr/FangYKSD24}, person re-identification \cite{HAMBARDE2024104917}, face detection \cite{DBLP:conf/fgr/MittalTMVS23}, biometric model compression \cite{DBLP:conf/biosig/NetoCCS23}, biometric sample quality \cite{DBLP:conf/icb/TerhorstKDKK20, DBLP:conf/biosig/GalballyHB18}, keystroke dynamics \cite{DBLP:journals/access/StragapedeVTMDFO24} or synthetic face data \cite{Huber_2024_WACVBias}.

\vspace{-4mm}
\subsection{Explainable Face Recognition}
\vspace{-2mm}
Explainability has received increasing attention in the field of FR \cite{DBLP:journals/corr/abs-2208-09500} because automated FR decisions are made that can have a major impact on our lives \cite{DBLP:journals/ijon/WangD21a}. Furthermore, privacy-related data is processed and biometric systems are often used in a security context (access control, identity checks) requiring the trust of the operator and user in the system \cite{DBLP:journals/tbbis/JainDE22, DBLP:journals/corr/abs-2208-09500}. This trust may not be sustainable if the behavior of the system is not understandable and possible causes of errors are not transparent. We will limit ourselves here to the most relevant works and for a more comprehensive overview, please refer to \cite{DBLP:journals/corr/abs-2208-09500}.

Some works in the area of biometrics \cite{DBLP:conf/iccv/ShiJ19, DBLP:conf/bmvc/HuberTKDK22, DBLP:conf/cvpr/NetoSCT23, 10595448} focused on the idea of increasing trust in the system by providing a decision uncertainty estimation. However, the majority of works followed the trend of computer vision and focused on visualizing important areas in images using heatmaps \cite{DBLP:conf/cvpr/WangWDYZDMH20, DBLP:conf/iccv/SelvarajuCDVPB17}. They are designed to be applied to classification problems and highlight areas in an image that lead to the prediction of a certain class. Approaches like GradCAM \cite{DBLP:conf/iccv/SelvarajuCDVPB17} or ScoreCAM \cite{DBLP:conf/cvpr/WangWDYZDMH20} cannot be applied to FR without changes, as state-of-the-art FR models are not designed and trained to learn a multi-class classification problem and as FR systems include a feature extraction and matching process. 

Therefore, tailored solutions have been proposed to tackle explainable FR systems. Lin \etal \cite{DBLP:journals/tomccap/LinLCWCH21} proposes xCos, a learnable module that provides a similarity map and an attention map, highlighting important areas in input face images for verification. Training-free approaches are provided by black-box approaches \cite{DBLP:conf/cvpr/Mery22, DBLP:conf/wacv/MeryM22, DBLP:conf/wacv/KnocheTHR23, DBLP:conf/wacv/LuXE24} which are based on the idea of manipulating the input image in the spatial domain while observing the change in model output. Based on the observed change in similarity or decision, similarity heatmaps are derived. A different approach is taken by white-box approaches which assume access to the model internals to utilize the model architecture, weights or gradients for the generation of their explanations. \cite{DBLP:conf/wacv/HuberLTD24} proposed xSSAB, which uses similarity score-based gradients to create positive and negative argument maps. Xu \etal \cite{10581925} proposed feature-guidance of the gradients to generate explanation maps. With the recent advances of large language models, textual explanations also gained attention \cite{DBLP:journals/access/DeAndresTameTVMFO24, DBLP:journals/ivc/CasconePP23}.


Most recently, in \cite{huber2024spatialexplanationsexplainableface}, FR explanations based on frequency have been proposed. Inspired by the observation that human imperceptible frequency patterns are utilized by CNNs \cite{DBLP:conf/cvpr/WangWHX20, DBLP:conf/cvpr/AbelloHW21}, they proposed to generate frequency heat plots (FHPs). These FHPs indicate which frequency bands are most influential on the similarity of two images as assigned by an FR model. In this work, we leverage this approach to explore the importance of different frequencies based on the ethnicity of samples.


\vspace{-5mm}
\section{Methodology}
\vspace{-4mm}
In this section, we describe our approach to explain bias in FR based on frequency patterns. We start by revisiting the approach of \cite{huber2024spatialexplanationsexplainableface} to create frequency-based explanations that provide importance scores for different frequency bands in a face verification process. Then we describe how we train FR models on unbalanced data to obtain models with larger performance disparities to amplify possible observations of bias. After that, we describe our explainability scheme in more detail. 


\vspace{-4mm}
\subsection{Frequency-based Explanations}
\label{meth}
\vspace{-2mm}
To obtain frequency explanations and importance scores for our analysis, we follow the approach proposed in \cite{huber2024spatialexplanationsexplainableface}. This frequency-based explainability method is applied to a typical FR setup including an FR model $N$ that extracts face representation of face images. In the approach, two face images $j, k$ of a pair, probe and reference image, are manipulated in the frequency domain before being passed to $N$. To be more precise, the images ($I^{S}$) are transformed into frequency domain ($I^{F}$) using Discrete Fourier Transform (DFT):
\footnotesize 
\begin{equation}
    I^{F} = DFT(I^{S}).
\end{equation}
Then, based on Euclidean distance \cite{huber2024spatialexplanationsexplainableface,DBLP:conf/cvpr/WangWHX20}, the frequency space is divided into disjoint frequency bands of the same size, e.g. frequencies 1 to 5 belong to the first frequency band. In the next step, the information in a certain frequency band ($b$) is removed by masking ($M$) and the images are transformed back to the spatial domain by applying the lossless Inverse Discrete Fourier transformation (iDFT):
\footnotesize
\begin{equation}
    I^{S}_{M} = iDFT(M_{b} \times I^{F}).
\end{equation}
The masking is done for each frequency band $b$, obtaining a set of image pairs in the spatial domain with different information in the frequency domain removed. The instances with different frequency bands masked are then processed by the FR model $N$ and the observed change in the similarity score $s_{c}$ compared with the similarity score of the unaltered image pair $s_{c}(N(j), N(k))$ is interpreted as the frequency band importance $h$:
\footnotesize
\begin{equation}
    h_{b} = | s_{c}(N(j), N(k)) - s_{c}(N(j_{M, b}), N(k_{M, b})) |
\end{equation}
For a higher interpretability \cite{huber2024spatialexplanationsexplainableface}, the obtained importance scores $h$ are normalized and scaled to sum up to 1 to create relative scores and allow a comparison of the importance between different pairs:
\footnotesize
\begin{equation}
    \hat{h_{b}} = \frac{h_{b}}{min(h_{b})},
\end{equation}
and
\footnotesize
\vspace{-2mm}
\begin{equation}
   \overline{h_{b}} = \frac{\hat{h_{b}}}{\sum(\hat{h_{b}})}.
\end{equation}
The frequency importance therefore describes the change in similarity score that is observed when the information present in the frequency is removed and therefore not considered by the FR model. For a more formalized and extensive description of the process, please refer to the original paper \cite{huber2024spatialexplanationsexplainableface}. 
A limitation of choosing this black-box masking approach as with all black-box pertubation-based approach \cite{DBLP:conf/cvpr/Mery22, DBLP:conf/wacv/MeryM22, DBLP:conf/wacv/KnocheTHR23, DBLP:conf/wacv/LuXE24} is, that out-of-distribution behavior can appear, since the model is not trained on (frequency-) masked data.

\vspace{-5mm}
\subsection{Amplifying the Ethnicity Bias of FR Models}
\vspace{-2mm}
As discussed in Section \ref{subsec:bias}, FR models are known to have performance variations across ethnicity groups \cite{DBLP:conf/ciarp/AcienMVBF18, DBLP:journals/tbbis/CavazosPCO21, DBLP:conf/cvpr/SVKAB19, DBLP:journals/tifs/KlareBKBJ12} without and even when applying approaches to mitigate this bias. To increase observable and analyzable results, we propose to intentionally increase the present ethnicity bias by training FR models on selected subsets that neglect certain ethnicity subgroups. Similar to \cite{10581908}, we assume that not being trained on a certain ethnicity subgroups reduce the performance of the model on the neglected data due to the model not being able to generalize well to this subgroup. This is done, along with presenting explanations of pre-trained state-of-the-art FR models that were not designed specifically with an amplified bias.

Beside the used pre-trained state-of-the-art models, we train five additional models in total. One reference model $M$, and four intentionally biased models $M_{\overline{bias}}$ ($M_{\overline{Afr}}$, $M_{\overline{Asi}}$, $M_{\overline{Cau}}$ and $M_{\overline{Ind}})$. The reference model $M$ is trained on all subsets of the utilized ethnicity-balanced dataset and is expected to be less ethnicity-biased in comparison to the other models. The other four models are trained on the ethnicity subsets with the respective dataset left out, e.g. the model $M_{\overline{Afr}}$ is trained on the $Asian$, $Caucasian$, and $Indian$ subset but not on the $African$ subset. That data imbalance can lead to bias in FR models has been shown by \cite{DBLP:journals/pami/WangZD22, DBLP:conf/iccv/WangDHTH19} and recent works investigating the explainability of bias in FR have also opted to use the same strategy to amplify the model bias to have clearer explanations \cite{10581908}.

\vspace{-4mm}
\subsection{Relative Frequency Ranking}
\vspace{-2mm}
The first approach we propose is to investigate the relative frequency importance between the different ethnicity subsets given an FR model. Given a set of importance values for different frequency bands as proposed in \cite{huber2024spatialexplanationsexplainableface}  that indicate the importance of the frequency bands for a decision, we assume that if the model is unbiased, there should be no clear pattern or differences between the frequency importance values based on the ethnicity. 

To discover possible patterns or differences based on ethnicity, we first calculate the mean frequency importance value $P$ for each frequency band $b$ and ethnicity $e$ on ethnicity-split testing data $d$:
\footnotesize
\vspace{-5mm}
\begin{equation}
    P_{b, e} = \frac{\sum_{i}^{O}{p_{i, b, e}}}{O},
\end{equation}
where ${p_{i, b, e}}$ refers to the frequency importance value for the frequency band $b$ of a sample $i$ of ethnicitiy $e$ and $O$ to the total number of importance values for frequency band $b$ of ethnicitiy $e$. Then we rank the obtained mean importance values $P_{b, e}$ for each frequency band $b$ to obtain the relative frequency ranking based on ethnicity $e$. This relative frequency ranking provides us with a visualization to investigate patterns, such as higher importance of lower or higher frequency bands for certain ethnicities $e$. The relative frequency ranking (categorical) rather than the nominal ranking, is motivated by the interest in the trend rather than the exact differences.

\vspace{-4mm}
\subsection{Impact of Bias on Frequency Importance}
\vspace{-2mm}
The second investigation aims at explaining how bias influences the importance of frequency bands for different demographic groups. We therefore investigate the question: Given different demographic groups, how does the relative importance of frequency bands change with the presence of bias? Are different frequency bands less or more important to an FR model when investigating a specific ethnicity if it is biased toward this ethnicity?

To do this, we follow the same approach as for the first investigation, using the mean importance values for each frequency band per ethnicity on ethnicity-split testing data for a model. But instead of only investigating one model, we compare the differences in frequency importance based on the models. To visualize this, we plot the mean frequency band importance distribution for each of the models and ethnicities. Utilizing the biased models we trained on unbalanced data that have an amplified bias towards a certain demographic group, we also explore how the answer to the question may change if the model bias regarding a demographic group is amplified.






\vspace{-4mm}
\section{Experimental Setup}
\vspace{-3mm}
\subsection{Training and Testing Dataset}
\vspace{-3mm}
As the evaluation dataset, we utilized the Racial Faces-in-the-Wild dataset (RFW) \cite{DBLP:conf/iccv/WangDHTH19}. The RFW dataset provides four ethnicity-split subsets for bias evaluation. The included ethnicities are: \textit{Asian}, \textit{African}, \textit{Indian}, and \textit{Caucasian}. The protocol provides in total of 24k pre-defined pairs. For each subset 6,000 pairs with 3,000 genuine (same identity) and 3,000 imposter (different identity) are defined. The pairs are intra-ethnic (Asian-Asian, African-African, ...) and the verification performance is reported in accuracy (in \%) for each intra-ethnic pair list. Additional experiments on the BFW data \cite{DBLP:conf/cvpr/RobinsonLHQ0T20} are provided in the supplementary material.

For training models on unbalanced data that amplify bias, we utilized the BUPT-Balanceface dataset \cite{DBLP:journals/pami/WangZD22, DBLP:conf/cvpr/WangD20}. This dataset provides ethnicity-based training subsets, covering the ethnicities $White/Caucasian$, $Indian$, $African$, and $Asian$. Each of these subsets contains around 300k images of 7k different identities. Since we always leave out one of the identity subsets, our four intentionally biased models are trained on 21k identities and around 900k images. The baseline model $M$ is trained on all 1.2M images of 28k identities. For the rest of the paper, we refer to the models trained on unbalanced data as biased models. 

\vspace{-4mm}
\subsection{Face Recognition Models}
\vspace{-2mm}
We trained five instances of biased models ($M$, $M_{\overline{Afr}}$, $M_{\overline{Asi}}$, $M_{\overline{Cau}}$ and $M_{\overline{Ind}}$), using the ResNet-34 architecture \cite{DBLP:conf/cvpr/HeZRS16} and utilize the state-of-the-art ElasticFace-Arc \cite{DBLP:conf/cvpr/BoutrosDKK22} loss. We set the hyper-parameters as recommended by \cite{DBLP:conf/cvpr/BoutrosDKK22} and use scale $s=64$, margin $m=0.5$, and standard deviation $\sigma=0.05$. We utilize a batch size of 128 and use Stochastic Gradient Descent (SGD) optimizer with an initial learning rate of 1e-1 \cite{DBLP:conf/cvpr/BoutrosDKK22}. The momentum is set to 0.9 and the weight decay to 5e-4. The learning rate is divided by 10 at 80k, 140k, 210k, and 280k training iterations. The total number of training epochs is 50. During the training, we apply random horizontal flipping with a probability of 0.5 \cite{DBLP:conf/cvpr/BoutrosDKK22}. The processed images are aligned and cropped to $112 \times 112 \times 3$ using MTCNN \cite{DBLP:journals/corr/ZhangZL016} following \cite{DBLP:conf/cvpr/DengGXZ19} and are normalized to have pixel values between -1 and 1. 

To broaden our investigation on a larger variety of network backbones, training datasets and loss functions, we also utilized established state-of-the-art pre-trained FR models. These models utilize the ResNet-50 architecture \cite{DBLP:conf/cvpr/HeZRS16} and were trained on CASIA-WebFace \cite{DBLP:journals/corr/YiLLL14a}, which is not an ethnicity-balanced dataset. We select these models specifically since they were trained on CASIA-WebFace which does not share identities with the RFW evaluation benchmark in contrast to the other commonly used FR training dataset MS1M \cite{DBLP:conf/eccv/GuoZHHG16}. 
The selected pre-trained models utilize the AdaFace \cite{DBLP:conf/cvpr/Kim0L22} loss and ElasticFace-Cos \cite{DBLP:conf/cvpr/BoutrosDKK22} loss and were obtained from the official respective authors. 

\vspace{-4mm}
\subsection{Bias Evaluation Metrics}
\vspace{-2mm}
To measure the bias of the models, we follow recent related works and report different metrics. We report the mean accuracy over the different ethnicities, the standard deviation (STD) \cite{DBLP:conf/cvpr/HuangWLDSWZZ23, DBLP:conf/cvpr/WangD20} and the Skewed Error Ratio (SER) \cite{DBLP:conf/cvpr/WangD20}. The SER is defined as \cite{DBLP:conf/cvpr/WangD20}: 
\footnotesize
\vspace{-4mm}
\begin{equation}
    SER = \frac{max_{g}Err(g)}{min_{g}Err(g)}
\end{equation}
where $g \in $ ethnicities. It is the ratio of the highest error rate to the lowest error rate among the different ethnicities. 

\vspace{-4mm}
\subsection{Frequency Explanations Parameters}
\vspace{-2mm}
To obtain the frequency band importance for the different models and based on different ethnicities, we utilize the frequency-based explanation method proposed in \cite{huber2024spatialexplanationsexplainableface}. For the frequency band size, we chose $s=4$ to get fine-grained importance values, while obtaining robust results \cite{huber2024spatialexplanationsexplainableface}. As the distance function in the frequency domain, we follow \cite{huber2024spatialexplanationsexplainableface} and utilize Euclidean distance to divide the frequencies in disjoint frequency bands.

\vspace{-4mm}
\section{Results}
\vspace{-4mm}
First, we report the findings of our investigation by providing the verification accuracy and the observed bias in the baseline and the biased models trained on unbalanced data, as well as the state-of-the-art pre-trained models. Then, we report and evaluate the relative frequency ranking for the different ethnicities. After that, we provide the mean frequency importance distribution for the baseline model, as well as the four biased models, and discuss observed differences in the frequency importance. Additional results on the BFW \cite{DBLP:conf/cvpr/RobinsonLHQ0T20} dataset are provided in the supplementary material.

\vspace{-5mm}
\subsection{Bias of FR Models}
\vspace{-2mm}
The analysis of the bias of the five trained models, as well as the two pre-trained models, is provided in Table \ref{tab:bias}. The highest mean accuracy (93.98\%) has been achieved, as expected, by our baseline model $M$ which was trained on the whole ethnicity-balanced BUPT-Balanceface dataset \cite{DBLP:journals/pami/WangZD22}. The models trained on unbalanced datasets achieved lower verification accuracies in comparison to the baseline model. This degradation in accuracy is clearly observed when the models are evaluated on the ethnicity subsets that were not part of their training dataset. For example, the model $M_{\overline{AFr}}$ trained without African faces dropped from 92.92 \% (baseline) to 80.25\% on African pairs. In terms of bias measured by STD and SER, the models trained without certain subsets show higher bias as an unbiased model would achieve an STD of 0 and SER of 1. Only model $M_{\overline{Cau}}$ trained without Caucasian faces decreased its bias which contributed to a major drop in performance on Caucasian pairs in comparison to the baseline model $M$. 

The pre-trained models, AdaFace \cite{DBLP:conf/cvpr/Kim0L22} and ElasticFace-Cos (EF-Cos) \cite{DBLP:conf/cvpr/BoutrosDKK22} trained on the unbalanced, widely used CASIA-WebFace \cite{DBLP:journals/corr/YiLLL14a} dataset also show biased behavior. 

\vspace{-6mm}
\begin{table*}[]
\centering
\begin{tabular}{c|c|c|c|c|c|ccc}
&   & \multicolumn{4}{|c|}{RFW}  & \multicolumn{3}{c}{Bias}        \\
Model & \multicolumn{1}{c|}{Training Data} & \multicolumn{1}{c}{African} & \multicolumn{1}{c}{Asian} &  \multicolumn{1}{c}{Caucasian} & \multicolumn{1}{c|}{Indian} & \multicolumn{1}{c}{Mean} & \multicolumn{1}{c}{STD}& \multicolumn{1}{c}{SER} \\ \hline \hline
$M$  & BUPT \cite{DBLP:journals/pami/WangZD22} & 92.92  & 93.30 & 95.67  & 94.02  &  93.98 &  1.05& 1.64 \\ \hline
$M_{\overline{Afr}}$ & w/o African & 80.25  & 92.33  & 94.88  & 93.15  & 90.15 & 5.79 & 3.86\\
$M_{\overline{Asi}}$  & w/o Asian & 92.30 & 83.97  & 95.22  & 93.32 & 91.20 & 4.31 & 3.35\\
$M_{\overline{Cau}}$  & w/o Caucasian  & 91.93 &  92.78 &  90.88  & 93.10 & 92.17 & 0.86 & 1.32\\
$M_{\overline{Ind}}$  & w/o Indian & 92.05 & 92.73 &  95.28 & 90.17  & 92.56 & 1.83 & 2.08\\ \hline  
AdaFace \cite{DBLP:conf/cvpr/Kim0L22} & CASIA-WebFace \cite{DBLP:journals/corr/YiLLL14a} & 85.77 & 84.95 & 93.25 & 87.85 & 87.96 & 3.23 & 2.23\\
EF-Cos \cite{DBLP:conf/cvpr/BoutrosDKK22} & CASIA-WebFace \cite{DBLP:journals/corr/YiLLL14a} & 85.18 & 84.20 & 92.65 & 88.03 & 87.51 & 3.28 & 2.15\\ \hline  \hline  
\end{tabular}
\caption{Verification accuracy [in \%] and bias of the five trained FR models and the two pre-trained models on the RFW \cite{DBLP:conf/iccv/WangDHTH19} dataset. Removing an ethnicity subgroup from the training data drastically reduces the verification performance on this group and in most cases increased bias. The pre-trained models also show ethnicity bias with large performance variations based on the ethnicity.}
\label{tab:bias}
\vspace{-11mm}
\end{table*}
\vspace{-5mm}
\subsection{Relative Frequency Ranking}
\vspace{-2mm}
With the relative frequency ranking, we investigate, if clear patterns or differences can be observed in the importance of frequencies based on the ethnicities. We provide the relative frequency ranking plots in Figure \ref{fig:importance} for the pre-trained state-of-the-art models and the five trained models. The y-axis indicates the place in the ranking of the ethnicity-based mean importance and the x-axis shows the corresponding frequency band. The categorical ranking has been chosen as we are more interesting in identifying trends than the actual differences. We provide the plots with nominal ranking in the supplementary material. The ranking of the baseline model is provided in Figure \ref{fig:impBase}. The visualization of the ranking shows that for the African ethnicity, low frequencies have a higher importance and high frequencies have less importance when compared to the other ethnicities. Similar behavior can be observed on all ranking plots, except for the AdaFace \cite{DBLP:conf/cvpr/Kim0L22} ranking, where high frequencies show a higher importance for African compared to the other ethnicities. For the Asian importance ranking it can be observed that in most cases, high frequencies are important for the models and low frequencies are less important compared to the other frequencies, i.e., the value on the y-axis is decreased when the value on the x-axis is increased. Based on this analysis, we conclude that for different ethnicities, different frequencies are utilized by the model, answering the question of our first investigation that models focus on different frequencies based on the ethnicity may be linked to bias.

\vspace{-5mm}
\subsection{Impact of Bias on Frequency Importance}
\vspace{-3mm}

\begin{figure}
     \centering
         \begin{subfigure}[b]{0.35\textwidth}
         \centering
         \includegraphics[width=\textwidth]{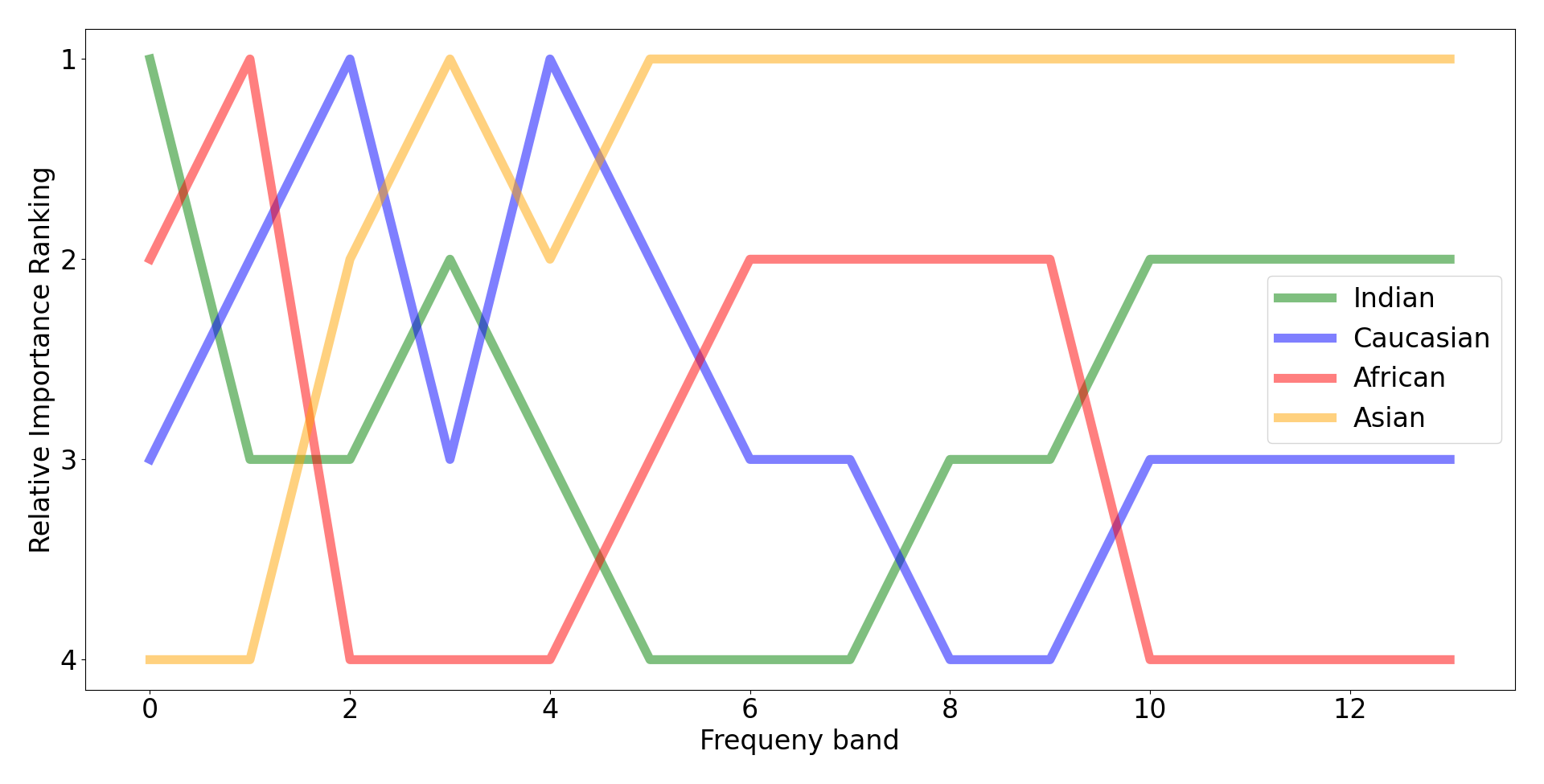}
         \caption{ElasticFace-Cos \cite{DBLP:conf/cvpr/BoutrosDKK22}}
         \label{fig:elasticfacecos}
     \end{subfigure}
     \begin{subfigure}[b]{0.35\textwidth}
         \centering
         \includegraphics[width=\textwidth]{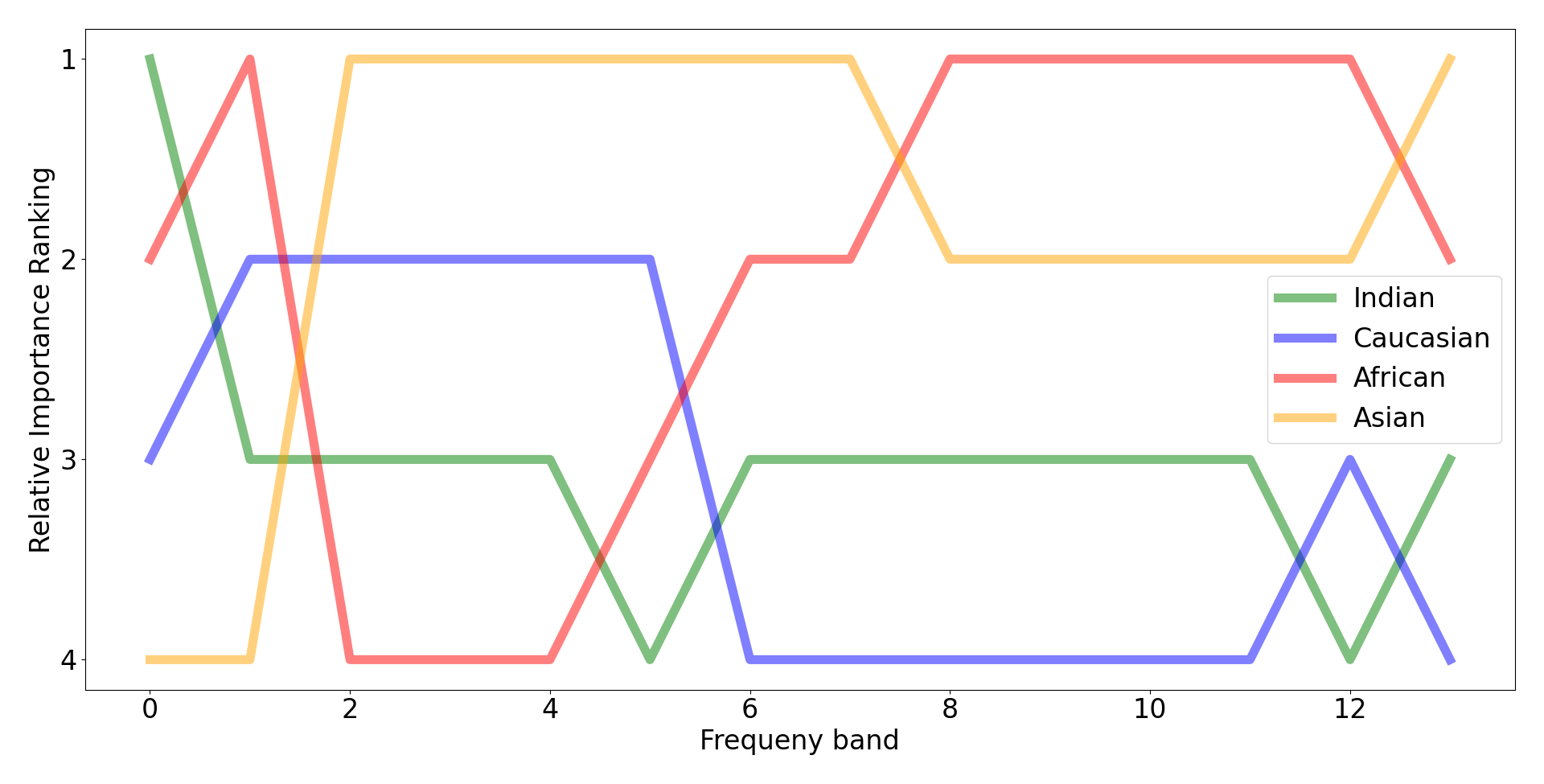}
         \caption{AdaFace \cite{DBLP:conf/cvpr/Kim0L22}}
         \label{fig:adaface}
         \end{subfigure}
     \begin{subfigure}[b]{0.35\textwidth}
         \centering
         \includegraphics[width=\textwidth]{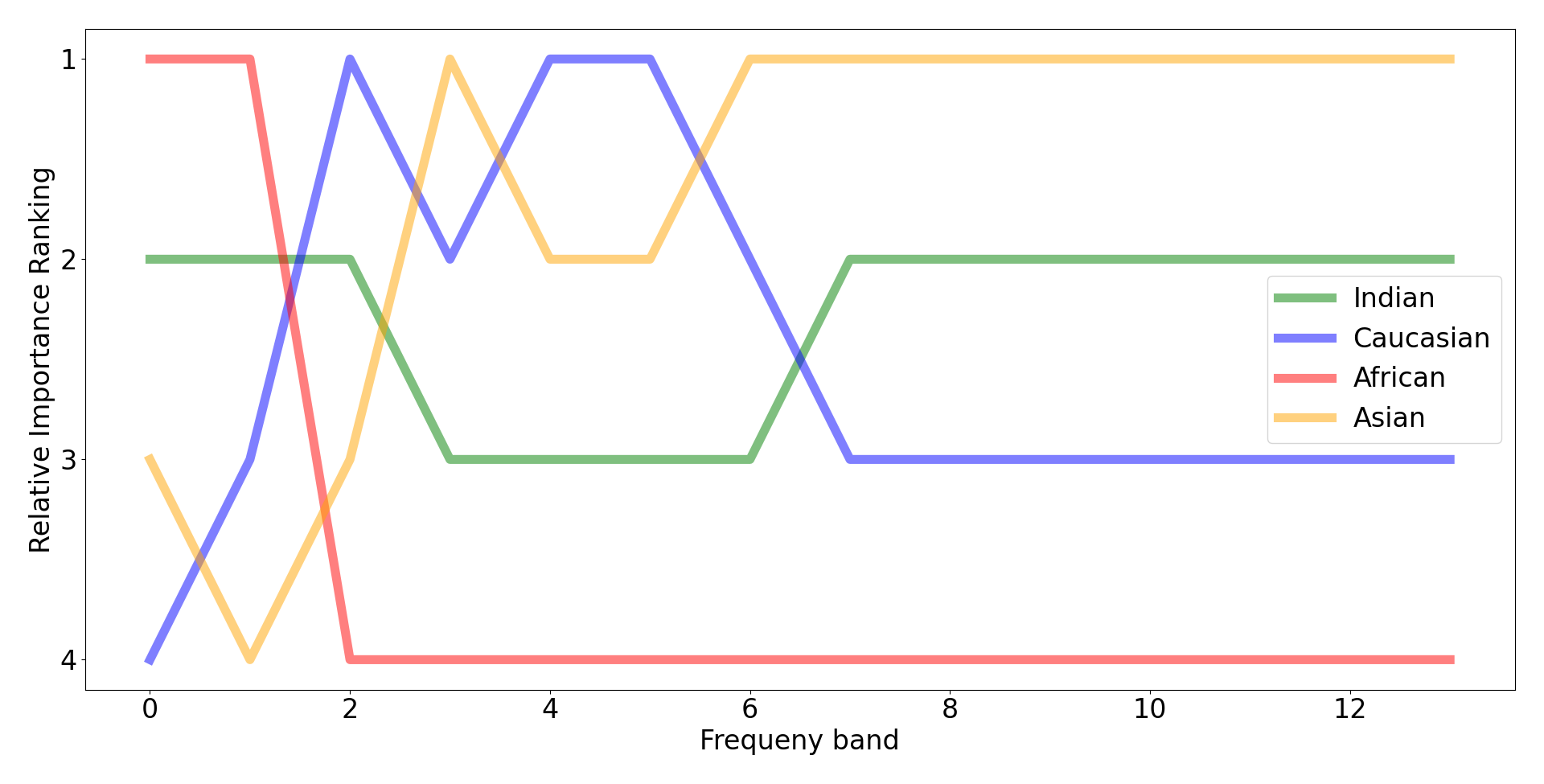}
         \caption{$M_{\overline{Afr}}$ (No African)}
         \label{fig:impNoAfrican}
     \end{subfigure}
     \begin{subfigure}[b]{0.35\textwidth}
         \centering
         \includegraphics[width=\textwidth]{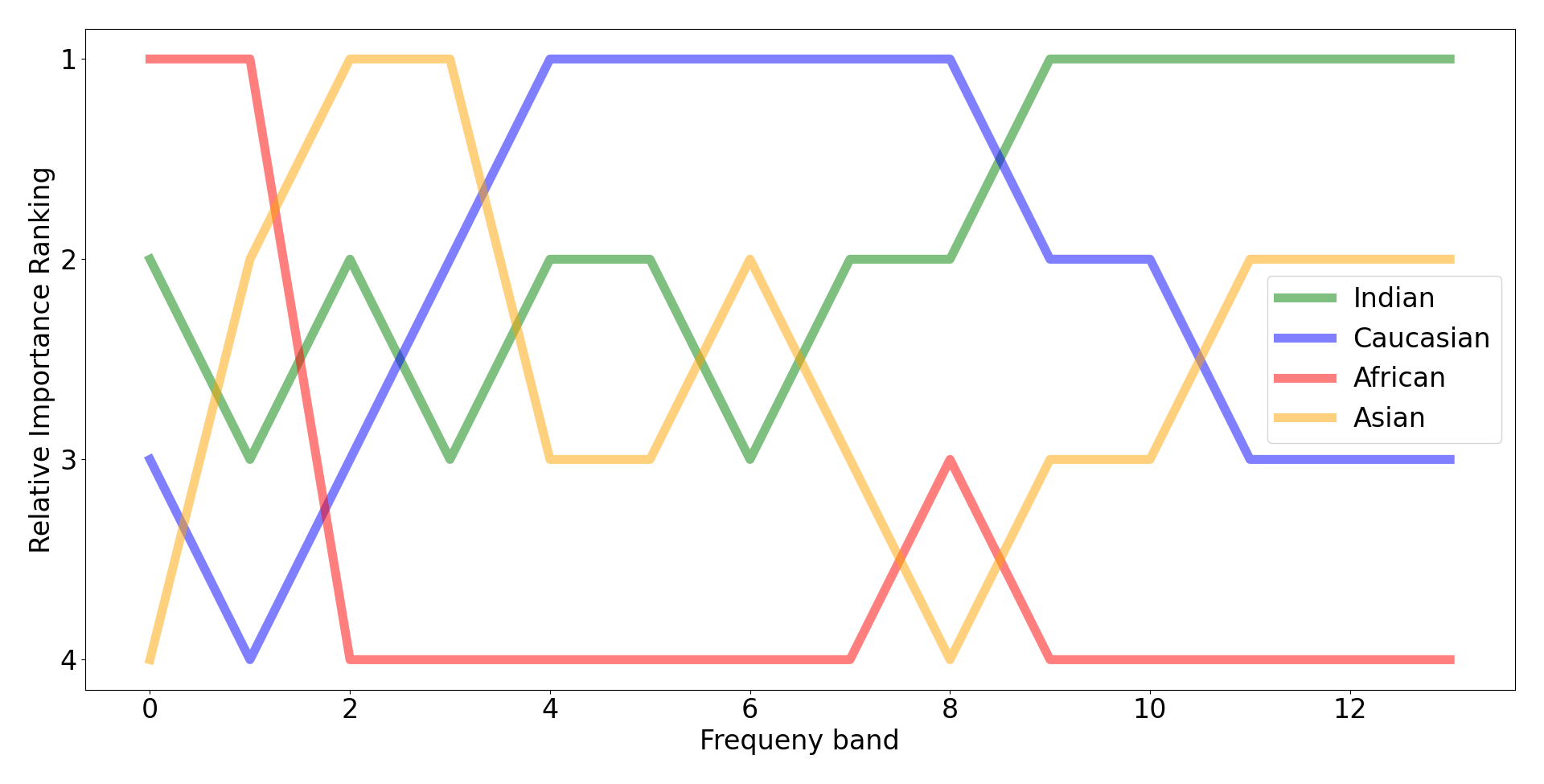}
         \caption{$M_{\overline{Asi}}$ (No Asian)}
         \label{fig:impNoAsian}
     \end{subfigure}
     \begin{subfigure}[b]{0.35\textwidth}
         \centering
         \includegraphics[width=\textwidth]{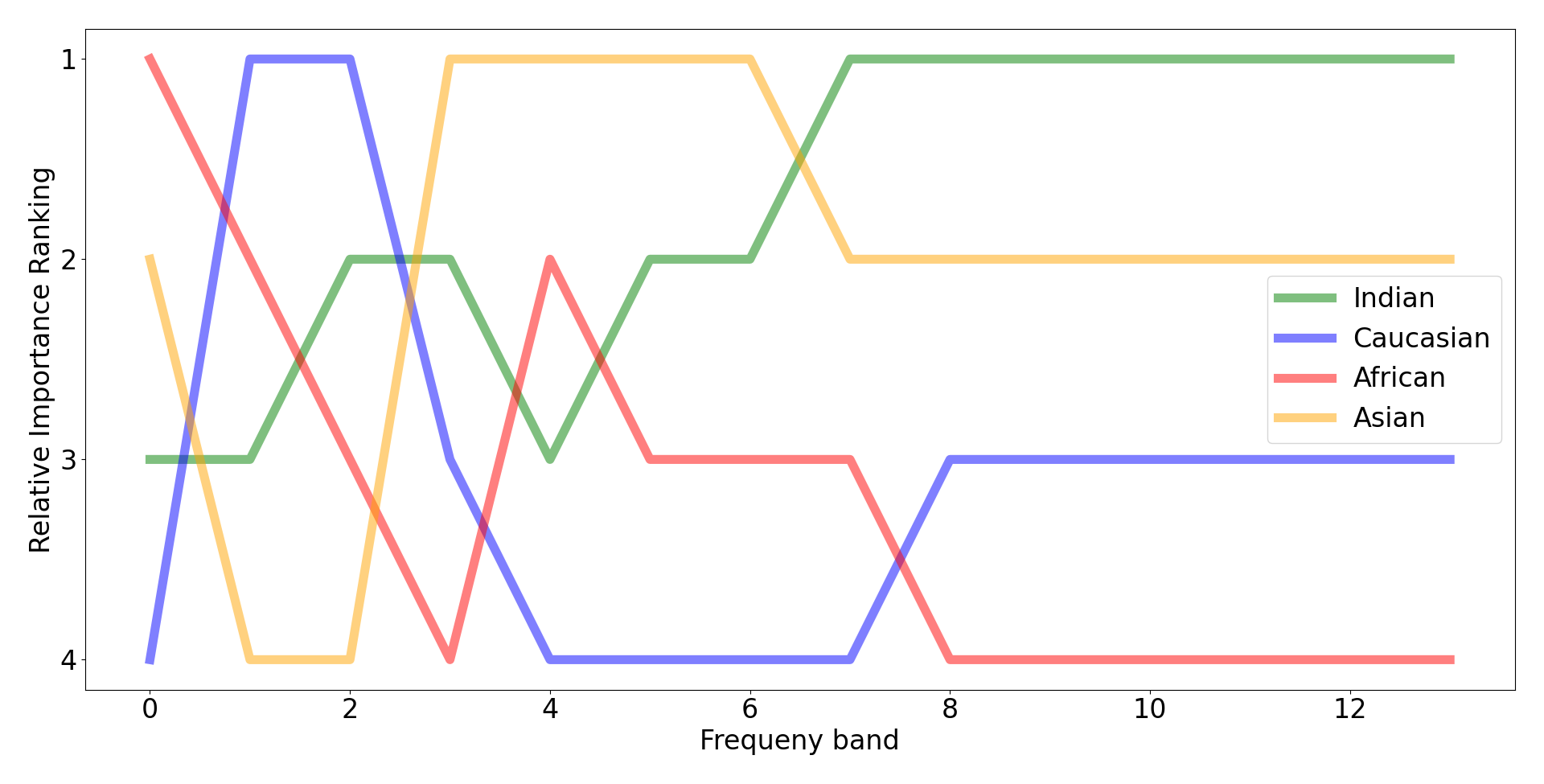}
         \caption{$M_{\overline{Cau}}$ (No Caucasian)}
         \label{fig:impNoCau}
     \end{subfigure}
     \begin{subfigure}[b]{0.35\textwidth}
         \centering
         \includegraphics[width=\textwidth]{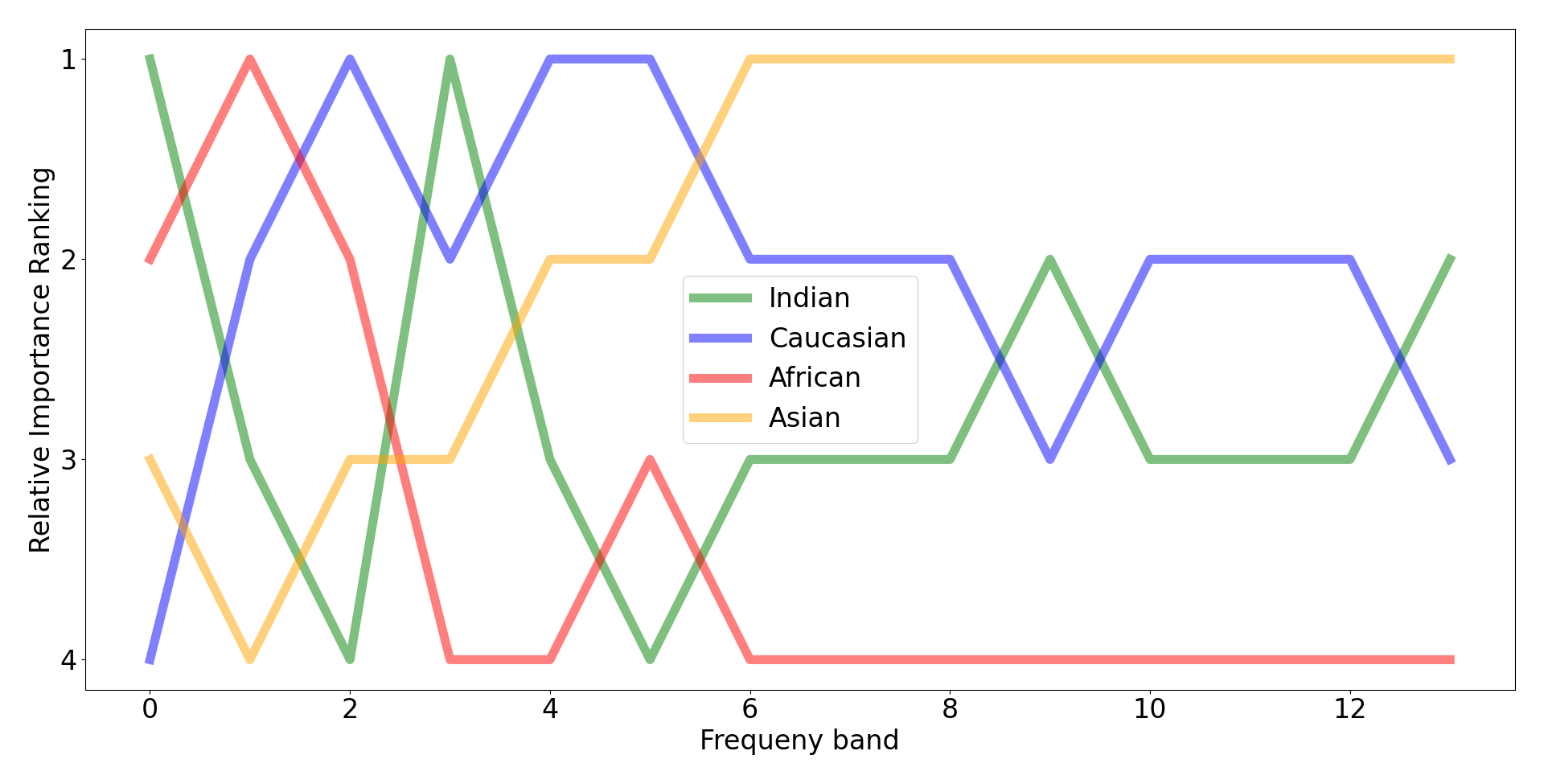}
         \caption{$M_{\overline{Ind}}$ (No Indian)}
         \label{fig:impNoInd}
     \end{subfigure}
     \begin{subfigure}[b]{0.35\textwidth}
         \centering
         \includegraphics[width=\textwidth]{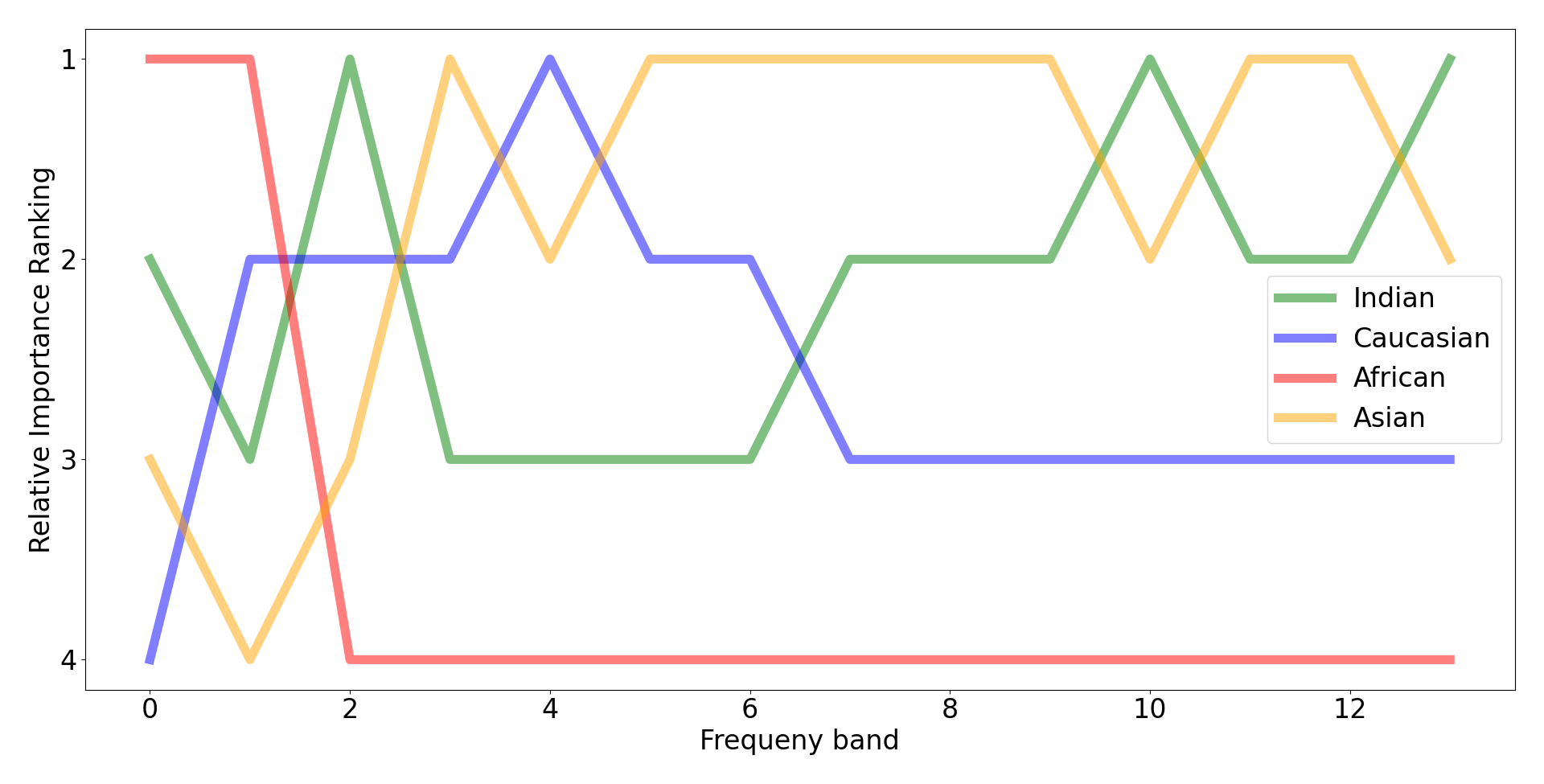}
         \caption{$M$ (Baseline)}
         \label{fig:impBase}
     \end{subfigure}
     \caption{Relative frequency ranking of the ethnicity for all models. Different pattern can be observed, such as low importance of high frequencies for African (red) or high importance of high frequencies for Asian (orange) in the most cases compared to the other ethnicities.}
    \label{fig:importance}
\end{figure}

\begin{figure}
\centering
\includegraphics[width=0.9\textwidth]{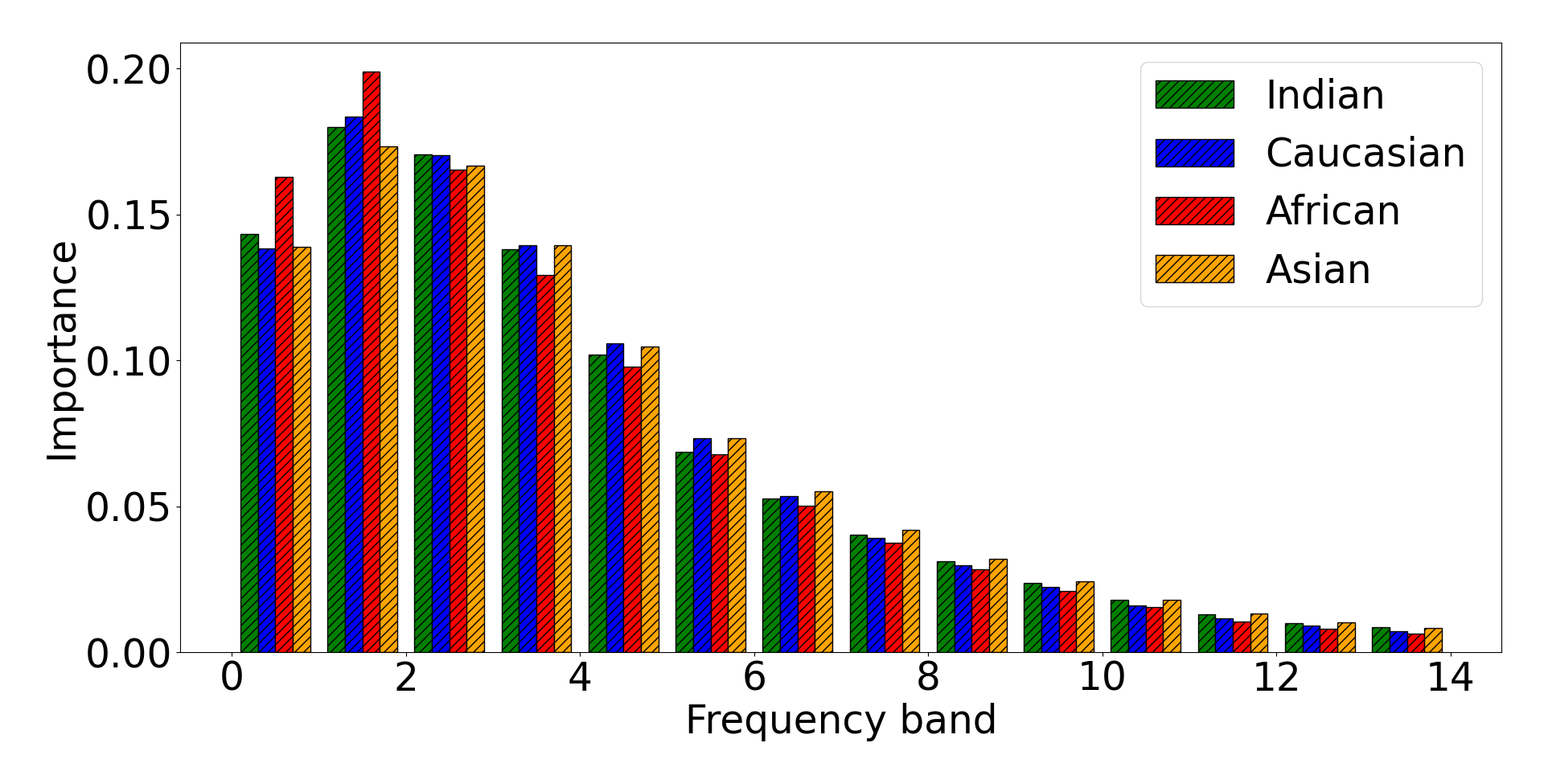}
         \caption{Mean frequency importance distribution based on ethnicity for model $M$}
         \label{fig:distBase}
\end{figure}

\begin{figure}
     \centering
     \begin{subfigure}[b]{\textwidth}
         \centering
         \includegraphics[width=0.9\textwidth]{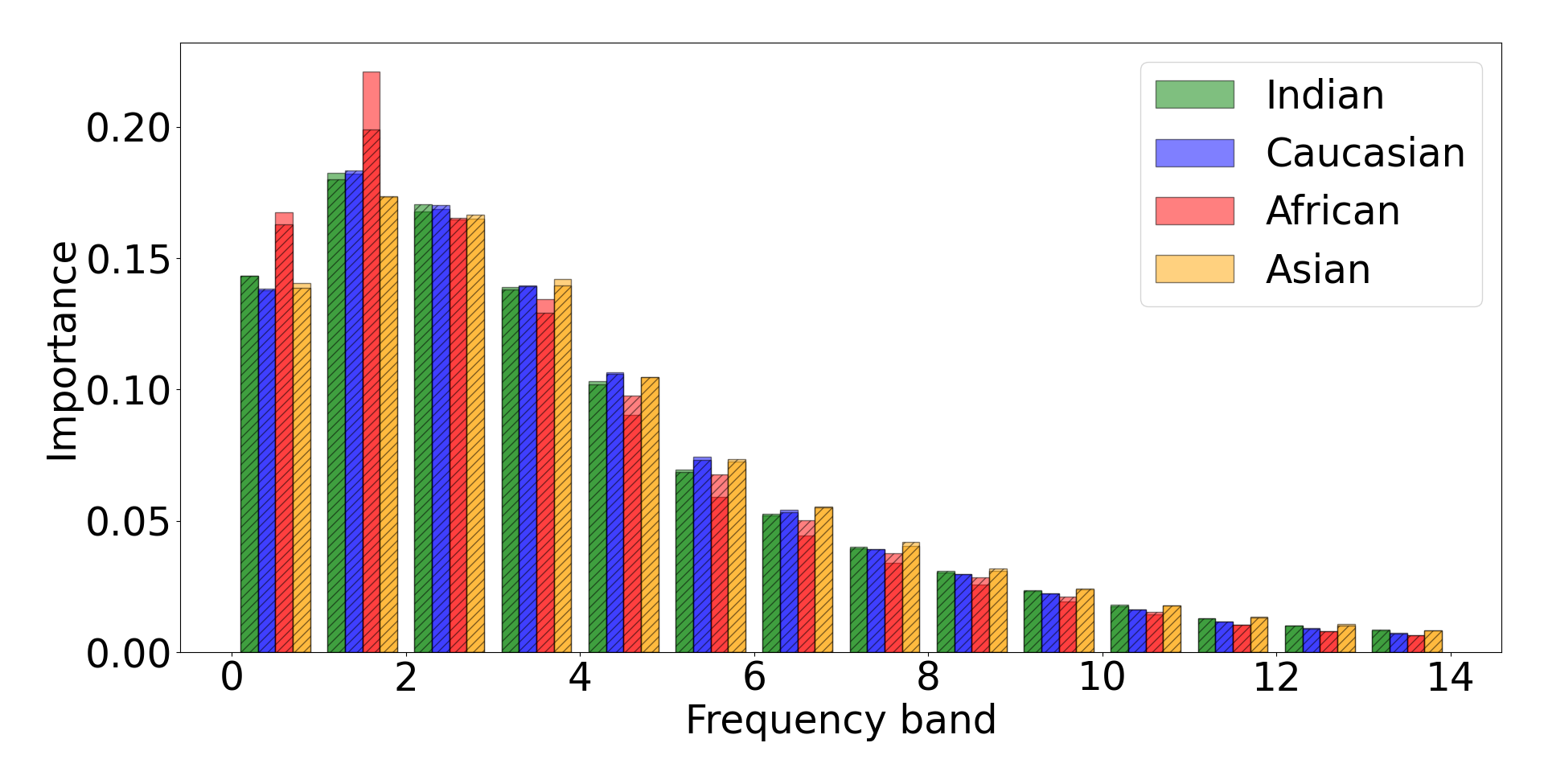}
         \caption{$M_{\overline{Afr}}$ (No African)}
         \label{fig:distNoAfrican}
     \end{subfigure}
     \begin{subfigure}[b]{\textwidth}
         \centering
         \includegraphics[width=0.9\textwidth]{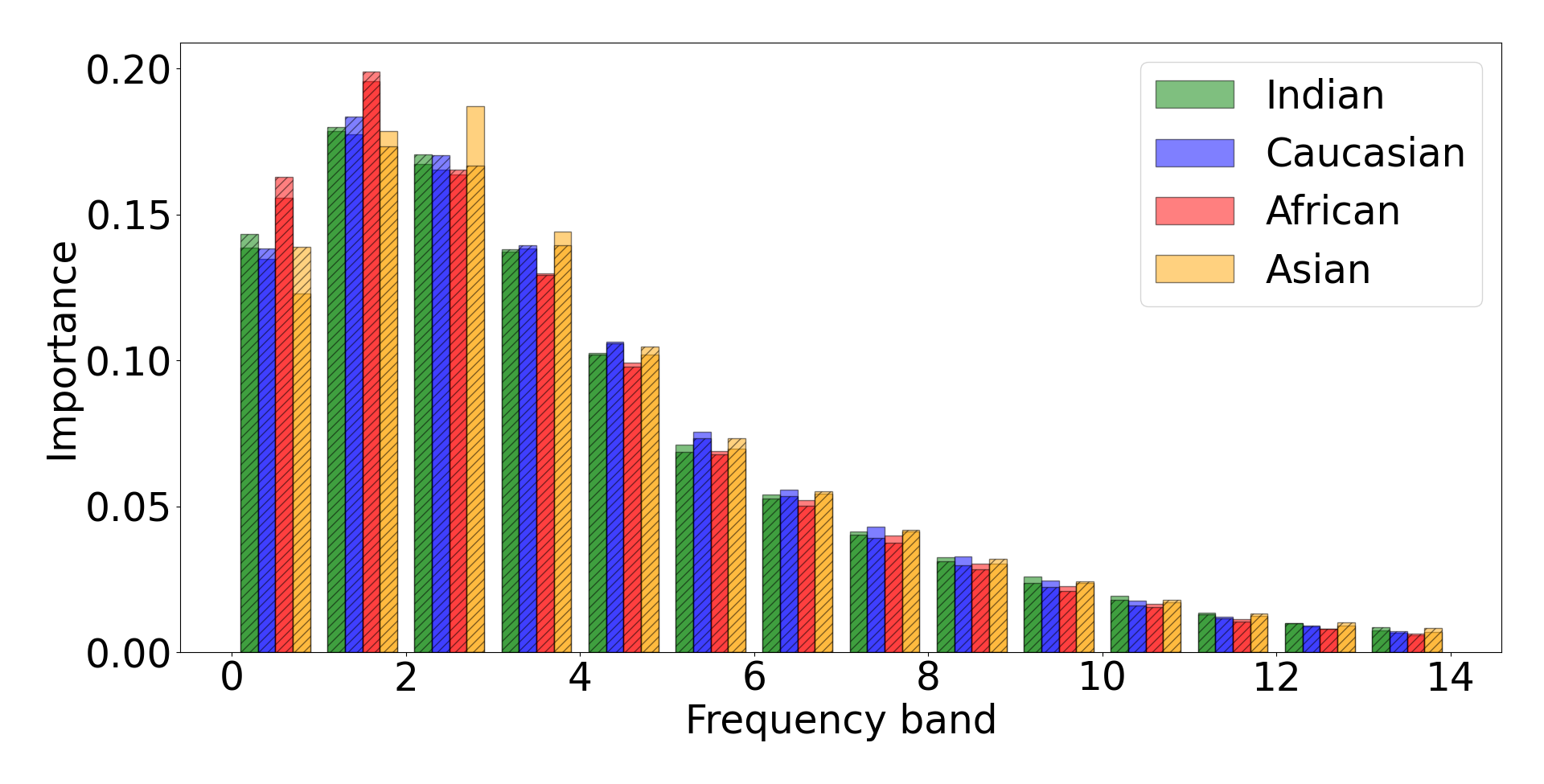}
         \caption{$M_{\overline{Asi}}$ (No Asian)}
         \label{fig:distNoAsian}
     \end{subfigure}
        \caption{Mean frequency importance distribution based on Ethnicity for model $M_{\overline{Afr}}$ and $M_{\overline{Asi}}$. The striped bar indicate the distribution of the baseline $M$, brightness changes indicate differences between $M$ and the biased models.}
        \label{fig:afrasi}
\end{figure}

\begin{figure}
     \centering
     \begin{subfigure}[b]{\textwidth}
         \centering
         \includegraphics[width=0.9\textwidth]{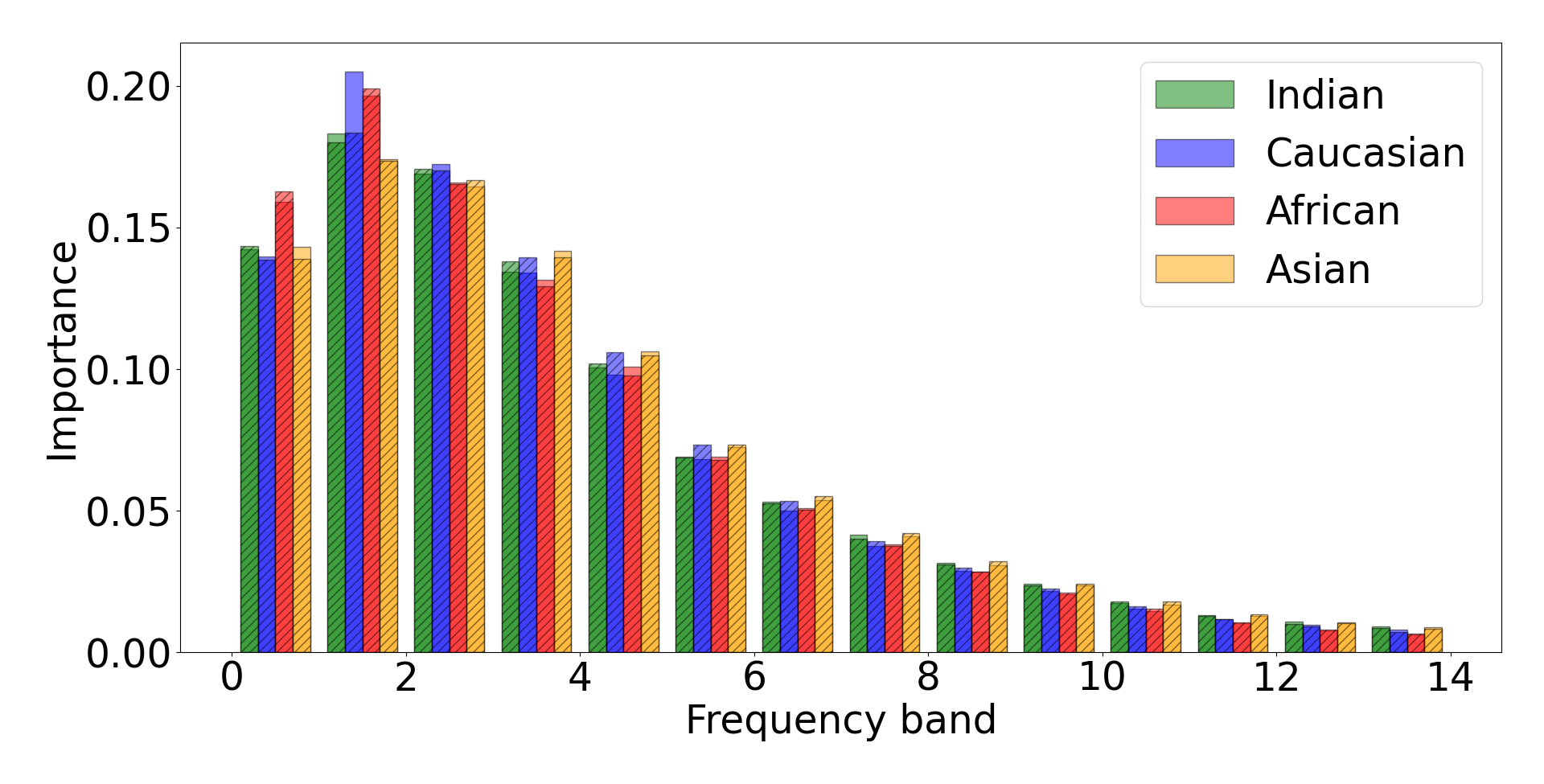}
         \caption{$M_{\overline{Cau}}$ (No Caucasian)}
         \label{fig:distNoCau}
     \end{subfigure}
     \begin{subfigure}[b]{\textwidth}
         \centering
         \includegraphics[width=0.9\textwidth]{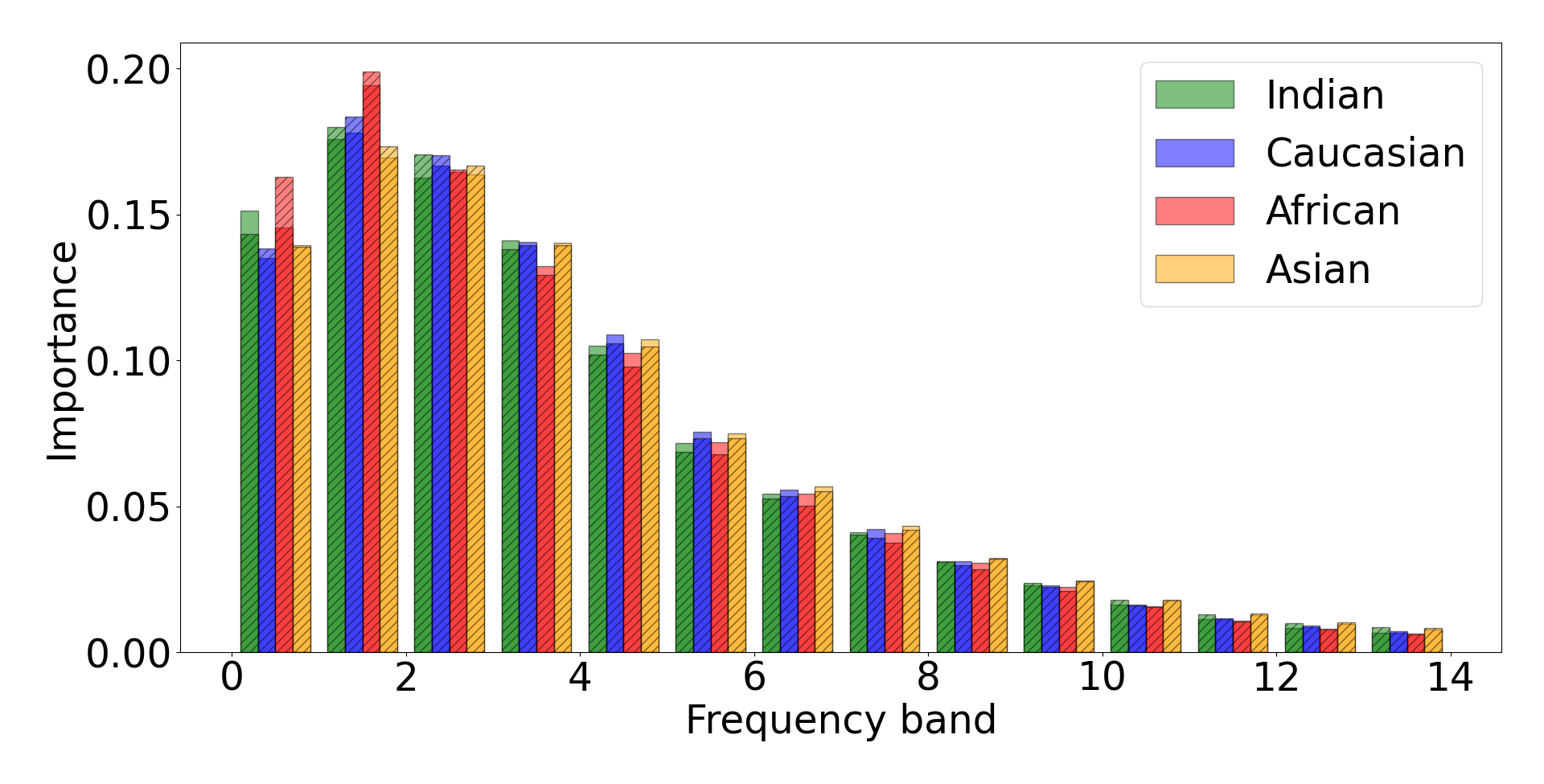}
         \caption{$M_{\overline{Ind}}$ (No Indian)}
         \label{fig:distNoInd}
     \end{subfigure}
        \caption{Mean frequency importance distribution based on Ethnicity for model $M_{\overline{Cau}}$ and $M_{\overline{Ind}}$. The striped bar indicate the distribution of the baseline $M$, brightness changes indicate differences between $M$ and the biased models.}
        \label{fig:cauind}
\end{figure}

\begin{figure}
     \centering
     \begin{subfigure}[b]{\textwidth}
         \centering
         \includegraphics[width=0.9\textwidth]{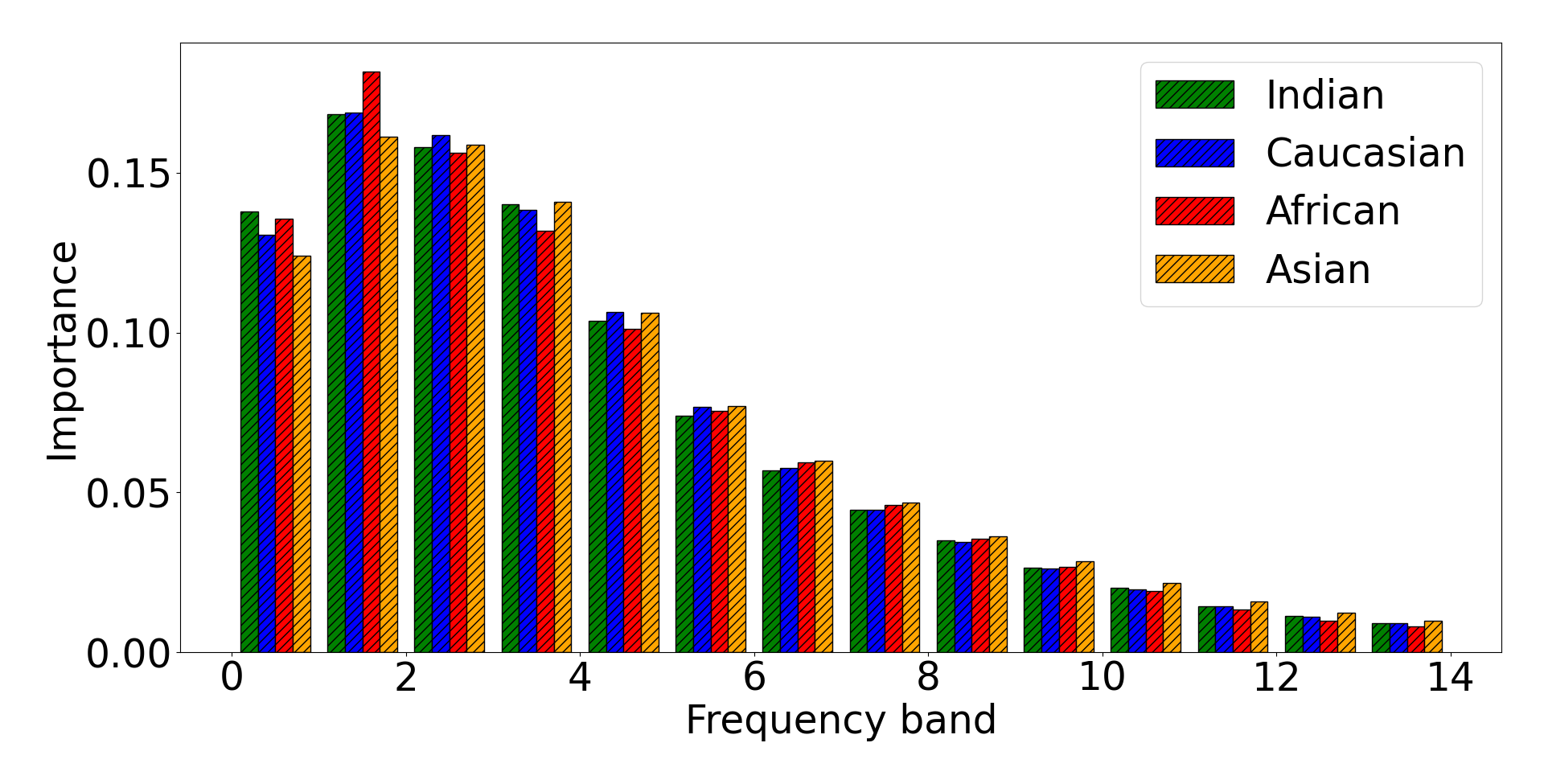}
         \caption{ElasticFace-Cos \cite{DBLP:conf/cvpr/BoutrosDKK22}}
         \label{fig:distElastic}
     \end{subfigure}
     \begin{subfigure}[b]{\textwidth}
         \centering
         \includegraphics[width=0.9\textwidth]{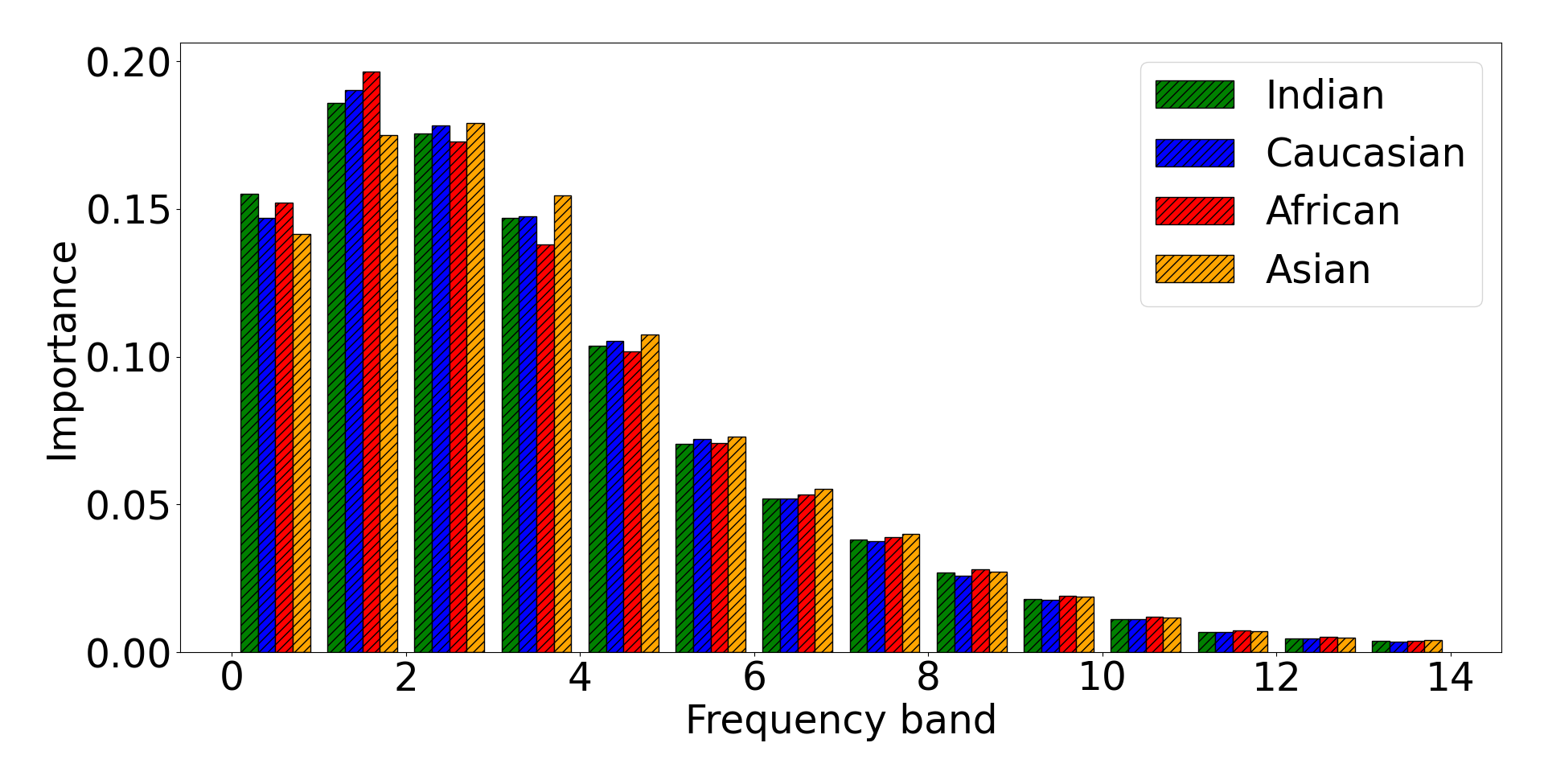}
         \caption{AdaFace \cite{DBLP:conf/cvpr/Kim0L22}}
         \label{fig:distAda}
     \end{subfigure}
        \caption{Mean frequency importance distribution based on ethnicity for the two state-of-the-art models.}
        \label{fig:}
\end{figure}

For different demographic groups, we investigate how the relative importance of frequencies changes with the presence of bias. We also provide the mean frequency importance distribution based on ethnicity for all considered models in Figure \ref{fig:distBase}, \ref{fig:distNoAfrican}, \ref{fig:distNoAsian}, \ref{fig:distNoCau}, \ref{fig:distNoInd}, \ref{fig:distElastic}, and \ref{fig:distAda}. For each investigated frequency band, the mean frequency importance based on ethnicity is plotted next to each other. On the y-axis we provide the mean frequency band importance and the x-axis shows the corresponding frequency band. In \ref{fig:distNoAfrican}, \ref{fig:distNoAsian}, \ref{fig:distNoCau}, and \ref{fig:distNoInd}, the distribution of $M$ is plotted for reference with striped bars to provide insights into the change of importance for each frequency band. The distribution of the biased models is also plotted but without bars. Bright areas with less colorfulness of the bars without stripes therefore indicate an increase in importance for the biased model, while bright areas with stripes indicate a decrease in importance. 
Since we focus here on the change of mean frequency importance, we additionally provide the standard deviation of the mean frequency importance plotted as error bars in the supplementary material.

In general, it can be observed that the most important frequencies for all the models are the lower frequencies, and with an increase in frequency, the frequencies get less important to the models. Furthermore, differences in the importance of certain frequency bands based on ethnicity (bars plotted next to each other) are shown, especially in the lower frequencies. This is also true for the best-performing baseline model $M$ (Figure \ref{fig:distBase}).

In the mean frequency importance distribution of $M$, it is shown that low frequencies are most important for African ethnicity, while higher frequencies are more important to Asian ethnicity. This is also represented in \ref{fig:impBase}. The largest difference exists in the importance of low frequencies (frequency bands 0 and 1).

When comparing the model $M_{\overline{Afr}}$ that is biased towards African (verification performance drop from 92.92\% (baseline $M$) to 80.25\% in Figure \ref{fig:distNoAfrican} with the baseline model, it shows that the importance of low frequencies for African is even higher (bright red area without stripes), while middle frequencies are less important (bright red areas with stripes). The difference between the African low frequency importance compared to the other ethnicity low frequency importance also increased. An increase in low frequency importance can also be observed in Figure \ref{fig:distNoCau} when investigating the mean frequency importance based on the Caucasian ethnicity. When investigating the Asian case (Figure \ref{fig:distNoAsian}) a larger difference in importance compared with the other ethnicities can be observed on the low frequency bands 0 and 2. While the importance of the lowest frequency band decreased, the importance of the 2 frequency band increased. In both cases, the difference with the other ethnicities increased. In Figure \ref{fig:distNoInd}, the importance of low frequencies when processing Indian increased for the $M_{\overline{Ind}}$, while the importance of low frequencies when processing African decreased. 

In \ref{fig:distElastic} and \ref{fig:distAda} the mean frequency importance distribution of the two state-of-the-art FR models is provided. It can be seen that there are also differences in the frequency importance based on ethnicity but it is less distinct when compared to the biased models.

Based on the investigated results discussed above, we observed that bias affects the mean frequency importance distribution of models. In most cases, the difference in frequency importance increased, especially in the low-frequency areas. When comparing the biased models with the baseline model, we observed major importance differences, especially in the low-frequency areas and less in the middle or high-frequency bands. Based on these observation we conclude that biased model tent to utilize low frequency patterns (higher importance) more than the baseline model. Our investigation compared between the overall mean difference between the same set of pairs across models. Given the large variation between pairs, another possibility is to look at the mean difference between individual pairs across these models.

\vspace{-5mm}
\vspace{-6mm}

\vspace{-3mm}
\section{Conclusion}
\vspace{-3mm}
In this work, we provided a new perspective on possible causes of ethnicity bias in deep learning-based FR models. Unlike previous works that focus on semantic and visual perspectives in explaining FR bias, we inspect bias in the frequency domain. This was motivated by recent studies pointing out that imperceptible frequency patterns by humans are perceptible and detectable by CNNs. As such patterns are utilized by CNNs, we applied a state-of-the-art frequency-based explanation method on two state-of-the-art models and trained five models with different ethnicity biases to explore bias in the frequency domain. Our analysis based on the impact of frequencies on the FR comparison scores showed, that different frequencies are more or less important to FR models based on ethnicity, and this effect is increased for more biased models. Possible future work includes investigating a possible link between reducing differences in frequency importance and reducing bias. Additionally,  investigations on different modalities and demographic attributes can be of interest.

\paragraph{Acknowledgement}
This research work has been funded by the German Federal Ministry of Education and Research and the Hessian Ministry of Higher Education, Research, Science, and the Arts within their joint support of the National Research Center for Applied Cybersecurity ATHENE. This work has been partially funded by the German Federal Ministry of Education and Research through the Software Campus Project.

\clearpage


%
%
\bibliographystyle{splncs04}
\bibliography{main}
\end{document}